\documentclass[11pt,a4paper,oneside]{article}
\usepackage[margin=0.7in]{geometry}
\usepackage[utf8]{inputenc}
\usepackage[T1]{fontenc}
\usepackage{amsmath,amssymb,amsfonts}
\usepackage{graphicx}
\usepackage{booktabs}
\usepackage{enumitem}
\usepackage{float}
\usepackage{caption}
\usepackage[dvipsnames]{xcolor}
\definecolor{citegreen}{RGB}{150,220,150}
\usepackage{colortbl}
\definecolor{lightgray}{gray}{0.85}
\usepackage{hanging}

\usepackage[
  colorlinks=false,          
  linkbordercolor=citegreen,
  citebordercolor=citegreen,
  urlbordercolor=cyan
]{hyperref}

\usepackage{array}
\usepackage{tcolorbox}

\setlist[itemize]{itemsep=1pt, topsep=2pt}
\setlist[enumerate]{itemsep=1pt, topsep=2pt}

\title{Specialists or Generalists? Multi-Agent and Single-Agent LLMs for Essay Grading}

\author{
    \textbf{Jamiu Adekunle Idowu}$^{1,2}$\thanks{Correspondence to: \href{mailto:jamiu@sahel.ai}{jamiu@sahel.ai}} \quad
    \textbf{Ahmed Almasoud}$^{3}$ \\
    \\
    {\small $^{1}$Sahel AI, Sahel Group Inc., Delaware, US} \\
    {\small $^{2}$University College London (UCL), United Kingdom} \\
    {\small $^{3}$Prince Sultan University, Saudi Arabia}
}
\date{}

\begin{document}
\maketitle

\begin{abstract}
Automated essay scoring (AES) systems increasingly rely on large language models, yet little is known about how architectural choices shape their performance across different essay quality levels. This paper evaluates single-agent and multi-agent LLM architectures for essay grading using the ASAP 2.0 corpus. Our multi-agent system decomposes grading into three specialist agents (Content, Structure, Language) coordinated by a Chairman Agent that implements rubric-aligned logic including veto rules and score capping. We test both architectures in zero-shot and few-shot conditions using GPT-5.1. Results show that the multi-agent system is significantly better at identifying weak essays while the single-agent system performs better on mid-range essays. Both architectures struggle with high-quality essays. Critically, few-shot calibration emerges as the dominant factor in system performance -- providing just two examples per score level improves QWK by approximately 26\% for both architectures. These findings suggest architectural choice should align with specific deployment priorities, with multi-agent AI particularly suited for diagnostic screening of at-risk students, while single-agent models provide a cost-effective solution for general assessment.
\end{abstract}

\section{Introduction}

The automated assessment of student writing is a long-standing challenge in educational technology, driven by the need to reduce the substantial workload teachers face in grading (\hyperlink{Sessler2025}{Seßler et al., 2025}; \hyperlink{Figueras2025}{Figueras et al., 2025}). While traditional methods have existed for decades, the emergence of Large Language Models (LLMs) has fundamentally altered the landscape of Automated Essay Scoring (AES). Unlike previous generations of automated graders, LLMs possess linguistic knowledge that allows them to analyze essays holistically and provide feedback that potentially benefits both teachers and students (\hyperlink{Mansour2024}{Mansour et al., 2024}). Recent studies have demonstrated that fine-tuning models like GPT-3.5 can significantly outperform earlier architectures such as BERT in scoring accuracy (\hyperlink{Latif2024}{Latif \& Zhai, 2024}), and that closed-source models continue to show advantages over open-source alternatives in alignment with human ratings (\hyperlink{Sessler2025}{Seßler et al., 2025}).

However, despite these advances, the optimal architectural deployment of LLMs for essay grading remains an open question. Proponents of multi-agent architectures argue that decomposing the grading task, separating rubric extraction, scoring, and feedback into distinct roles, can enhance performance, interpretability, and human-machine agreement (\hyperlink{WangDing2025}{Wang, Ding, et al., 2025}; \hyperlink{Su2025}{Su et al., 2025}). Conversely, other research indicates that increasing complexity does not guarantee better results. For instance, recent comparisons in student reflection assessment found that single-agent strategies utilizing few-shot prompting actually achieved higher match rates with human evaluators than multi-agent alternatives (\hyperlink{Li2025}{Li et al., 2025}). Similarly, in collaborative problem-solving contexts, multi-agent workflows failed to improve accuracy relative to single-agent models (\hyperlink{WangGopal2025}{Wang, Gopalakrishnan, \& Bergner, 2025}).

This divergence in findings highlights a critical gap in the literature, especially since little is known about how architectural choices interact with essay quality at a granular level. Does a multi-agent system justify its increased computational cost across the entire score distribution? This study addresses this gap by conducting a rigorous comparison of Single-Agent and Multi-Agent architectures using the ASAP 2.0 corpus (\hyperlink{Crossley2025}{Crossley et al., 2025}).

\section{Related Work}

\subsection{LLM Capabilities in Automated Scoring}
The application of LLMs to automated grading is gaining traction, with results suggesting they can effectively reduce teacher workload (\hyperlink{Sessler2025}{Seßler et al., 2025}). Comparisons between models, such as ChatGPT and Llama, indicate that performance is highly dependent on prompt engineering, yet these models generally exhibit comparable average performance to state-of-the-art methods while offering the added benefit of qualitative feedback (\hyperlink{Mansour2024}{Mansour et al., 2024}). However, challenges regarding consistency and bias remain. \hyperlink{Errica2025}{Errica et al. (2025)} highlighted the sensitivity of LLMs to minor prompt variations, proposing metrics to measure consistency across rephrasings. Furthermore, \hyperlink{Guo2025}{Guo et al. (2025)} investigated scoring disparities toward English Language Learners (ELLs), finding that while large training datasets can mitigate bias, limited samples may lead to scoring distortions. These findings underscore the need for robust evaluation frameworks when deploying LLMs for assessment.

\subsection{Multi-Agent Architectures in Education}
The integration of multi-agent systems represents a significant trend in educational AI. These systems typically decompose complex assessment tasks into specialized roles. Some frameworks introduced by recent studies include:

\begin{itemize}
    \item \textbf{AutoSCORE} (\hyperlink{WangDing2025}{Wang, Ding et al., 2025}) used two agents, one for component extraction and one for scoring, to ensure reasoning follows a structured process, reporting improvements over single-agent baselines.
    
    \item \textbf{CAFES} (\hyperlink{Su2025}{Su et al., 2025}) implemented a collaborative framework with a "Reflective Scorer" that iteratively refines scores based on feedback, achieving high agreement with ground truth in multimodal contexts.
    
    \item \textbf{CAELF} (\hyperlink{Hong2025}{Hong et al., 2025}) and \textbf{EvaAI} (\hyperlink{Lagakis2024}{Lagakis \& Demetriadis, 2024}) focused on the interactive and human-mimicking aspects of grading, using multiple agents to generate contestable feedback or simulate tutor-student interactions.
    
    \item \textbf{Decomposition for Specificity:} \hyperlink{Kang2026}{Kang and Kong (2026)} demonstrated the value of decomposition by breaking down "Topic Relevance" into sub-characteristics, allowing for more precise assessment in Chinese composition. Similarly, \hyperlink{Chu2024}{Chu et al. (2024)} introduced Rationale-based Multiple Trait Scoring (RMTS), where separate agents generate qualitative rationales to assist the quantitative scoring model.
\end{itemize}

\subsection{The Architectural Trade-off: Single vs. Multi-Agent}
Despite the proliferation of multi-agent frameworks, recent empirical evidence suggests that complexity does not always guarantee superior performance. While \hyperlink{WangDing2025}{Wang, Ding, et al. (2025)} found multi-agent designs superior for complex rubrics, \hyperlink{Li2025}{Li et al. (2025)} reported that a single-agent strategy with few-shot prompting outperformed multi-agent strategies in assessing student reflections. Furthermore, \hyperlink{WangGopal2025}{Wang, Gopalakrishnan, and Bergner (2025)} noted that while multi-agent systems could generate transparent rubrics, they did not improve classification accuracy for collaborative problem-solving items compared to single-agent models.

\section{Methodology}

The overall experimental workflow, comparing the Single-Agent and Multi-Agent pathways across different prompting conditions, is illustrated in Figure 1.

\begin{figure}[H]
    \centering
    \includegraphics[width=1\textwidth]{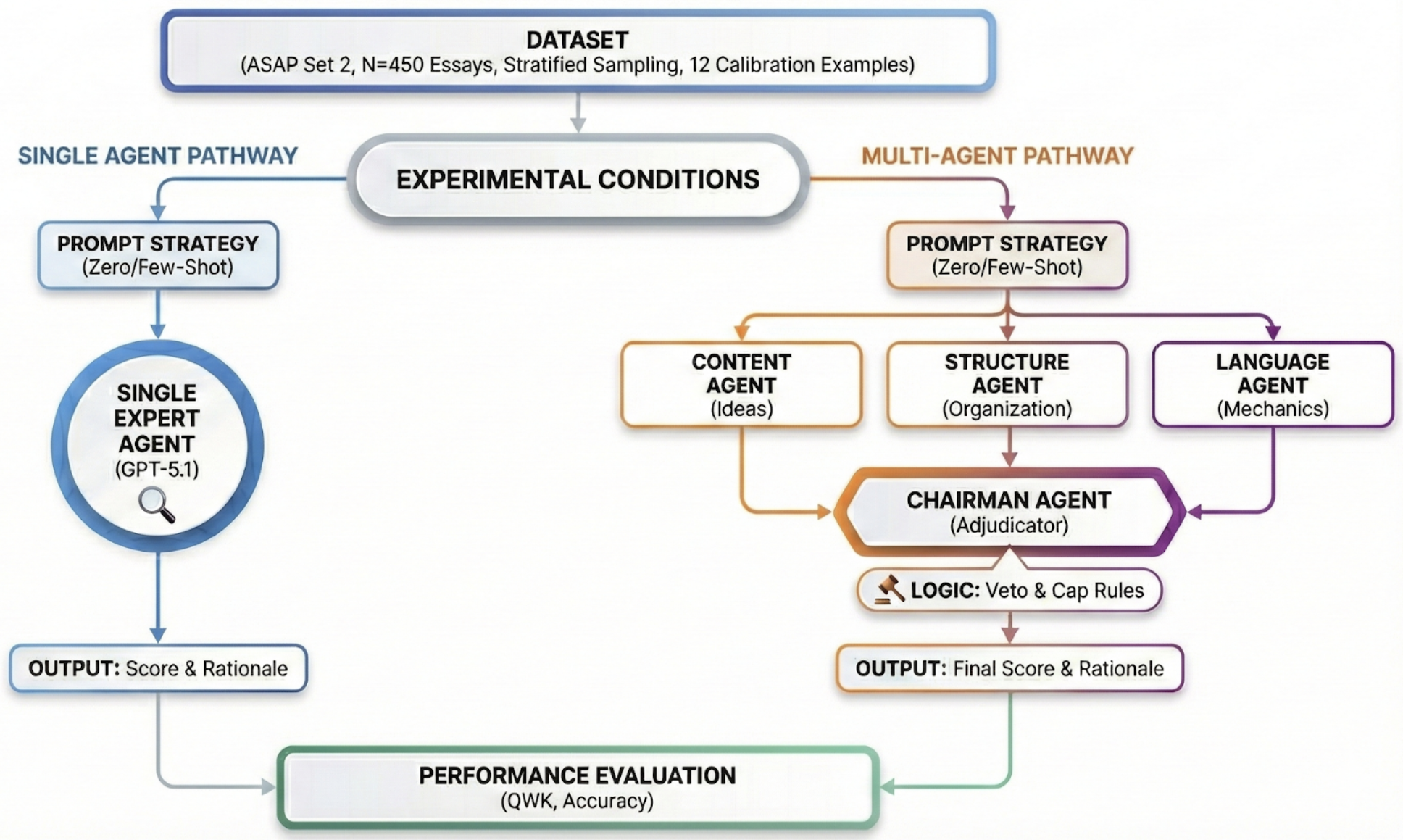} 
    \caption{High-level overview of the methodology}
    \label{fig:workflow}
\end{figure}

\subsection{Dataset}
This study uses data extracted from the ASAP 2.0 corpus -- designed to support automated essay scoring (AES) research (\hyperlink{Crossley2025}{Crossley et al., 2025}). The full ASAP 2.0 comprises 24,278 source-based argumentative essays written by U.S. secondary students (grades 6 - 10) across seven distinct prompts. \hyperlink{Crossley2025}{Crossley et al. (2025)} also provided an official rubric used by expert raters. The rubric, adapted from Scholastic Aptitude Test (SAT), used a 1–6 scale that had interval levels with a score of 6 indicating consistent mastery of writing. From this corpus, we employed stratified sampling to extract a representative test set of 450 essays. Additionally, for few-shot prompting, we curated a separate set of "calibration examples" (2 examples per score level, totaling 12 essays), ensuring no overlap between the calibration and test data. We also used the official full rubric in our prompts.

\subsection{Experiment}
We compared two distinct architectural approaches to automated grading: a Monolithic (Single-Agent) approach and a Hierarchical (Multi-Agent) approach. Both architectures were tested in Zero-Shot (rubric only) and Few-Shot (rubric + calibration examples) environments. All experiments were conducted using the GPT-5.1 model via the OpenAI API.

\subsubsection{Baseline: Single-Agent Architecture}
In the single-agent configuration, the model acted as the sole expert examiner. It was provided with the full rubric text and asked to output a holistic score (1–6) and a rationale.
\begin{itemize}
    \item \textbf{Zero-Shot:} The model relied solely on the full rubric.
    \item \textbf{Few-Shot:} The model was provided with 12 calibration examples (2 for each score level) to demonstrate the specific grading standards of the dataset.
\end{itemize}

\subsubsection{Multi-Agent Architecture}
Our multi-agent architecture consists of three Specialist Agents and one Adjudicator (Chairman Agent). We decomposed the grading task into three distinct roles with the specialist agents assuming one role each; and we used Negative Constraint Prompting (e.g., “Ignore grammar”) to ensure specialists remained within their domain. This decomposition strategy aligns with recent frameworks that deconstruct core assessment dimensions, such as topic relevance or specific traits, to improve the precision of LLM-based evaluations (\hyperlink{Kang2026}{Kang \& Kong, 2026}; \hyperlink{Chu2024}{Chu et al., 2024}).

\begin{itemize}
    \item \textbf{Content Agent:} Focused exclusively on ideas and evidence (instructed to ignore grammar).
    \item \textbf{Structure Agent:} Focused on organization and coherence (instructed to ignore facts or grammar).
    \item \textbf{Language Agent:} Focused on mechanics and vocabulary (instructed to ignore argument).
    \item \textbf{Chairman Agent (Adjudicator):} A final agent received the reports and scores from the three specialists. Unlike a simple averaging algorithm, the Chairman agent was prompted with specific logic-based heuristics (aligning with official rubric) to determine the final score. This includes the veto rule which states if any specialist agent assigned a score of 1, the final score by the Chairman Agent must be 1. It also includes the capping logic that if any specialist agent assigned a 2, the final score is capped at 2 or 3.
\end{itemize}

The structural composition of this system is shown in Figure 2, which visualizes the flow of information from the essay input through the three specialist agents to the Chairman Agent, where the specific veto and score-capping logic is applied.

\begin{figure}[H]
    \centering
    \includegraphics[width=1\textwidth]{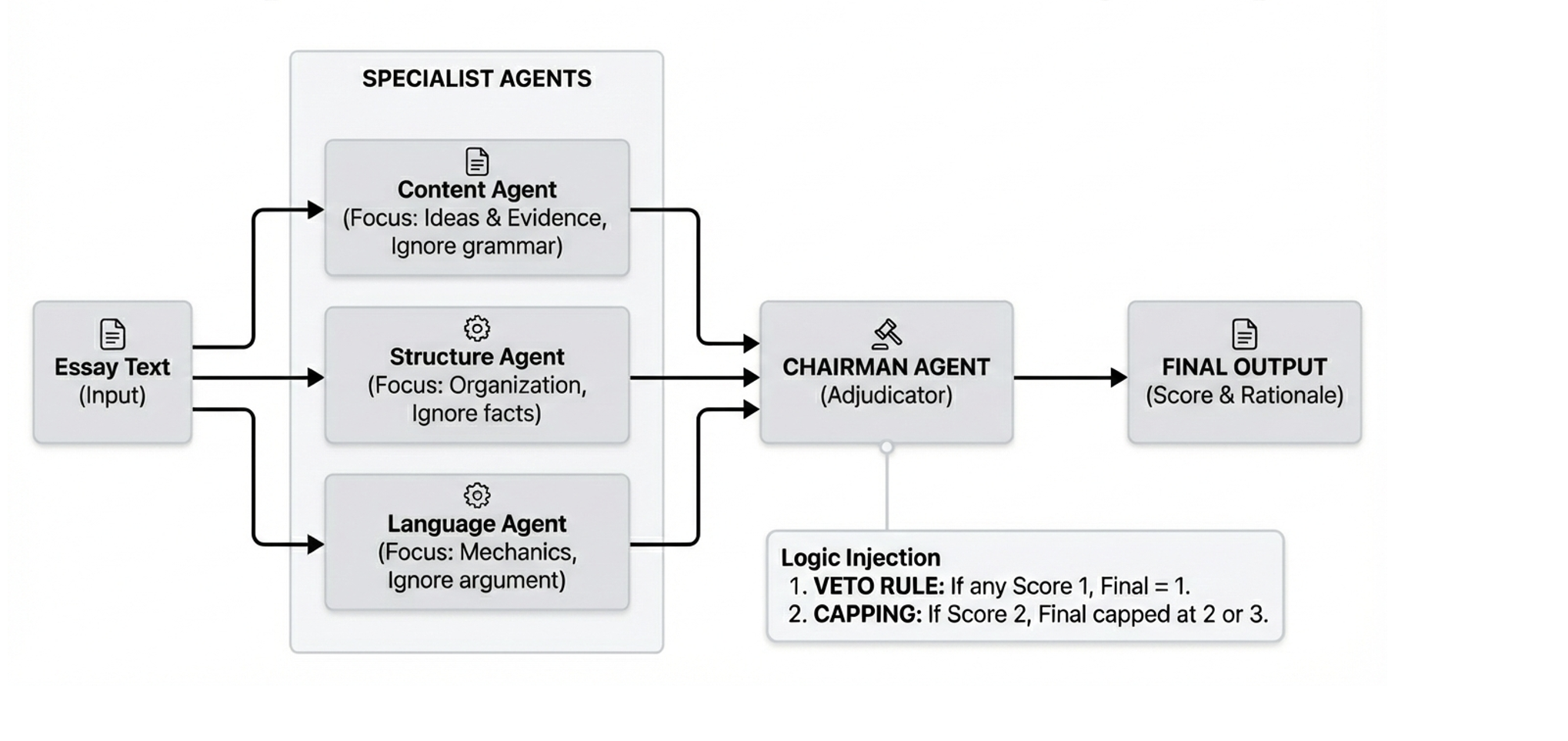} 
    \caption{The structural composition of the multi-agent system.}
    \label{fig:multiagent}
\end{figure}

\subsubsection{Evaluation}
Performance was evaluated against human scores (ground truth) using Quadratic Weighted Kappa (QWK), the primary metric for measuring agreement between raters while penalizing large discrepancies. We also reported Accuracy (Exact Match) i.e. percentage of predictions matching human scores exactly.

\section{Results and Discussion}

Table 1 presents the overall performance of the four experimental conditions tested in this study. The multi-agent architecture with few-shot prompting achieved the highest performance with a QWK of 0.7453 and an exact match accuracy of 57.11\%. This configuration outperformed the single-agent few-shot baseline by 2.88 percentage points in QWK (0.7165 vs. 0.7453) and 3.11 percentage points in exact match accuracy.

\begin{table}[H]
    \centering
    \caption{Overall Performance Metrics Across All Architectures}
    \begin{tabular}{lcc}
    \toprule
    Model & QWK & Exact Match \\
    \midrule
    Single-Agent (Zero) & 0.5664 & 49.11\% \\
    Single-Agent (Few) & 0.7165 & 54.00\% \\
    Multi-Agent (Zero) & 0.5917 & 47.56\% \\
    Multi-Agent (Few) & 0.7453 & 57.11\% \\
    \bottomrule
    \end{tabular}
\end{table}

\begin{figure}[H]
    \centering
    \includegraphics[width=0.65\linewidth]{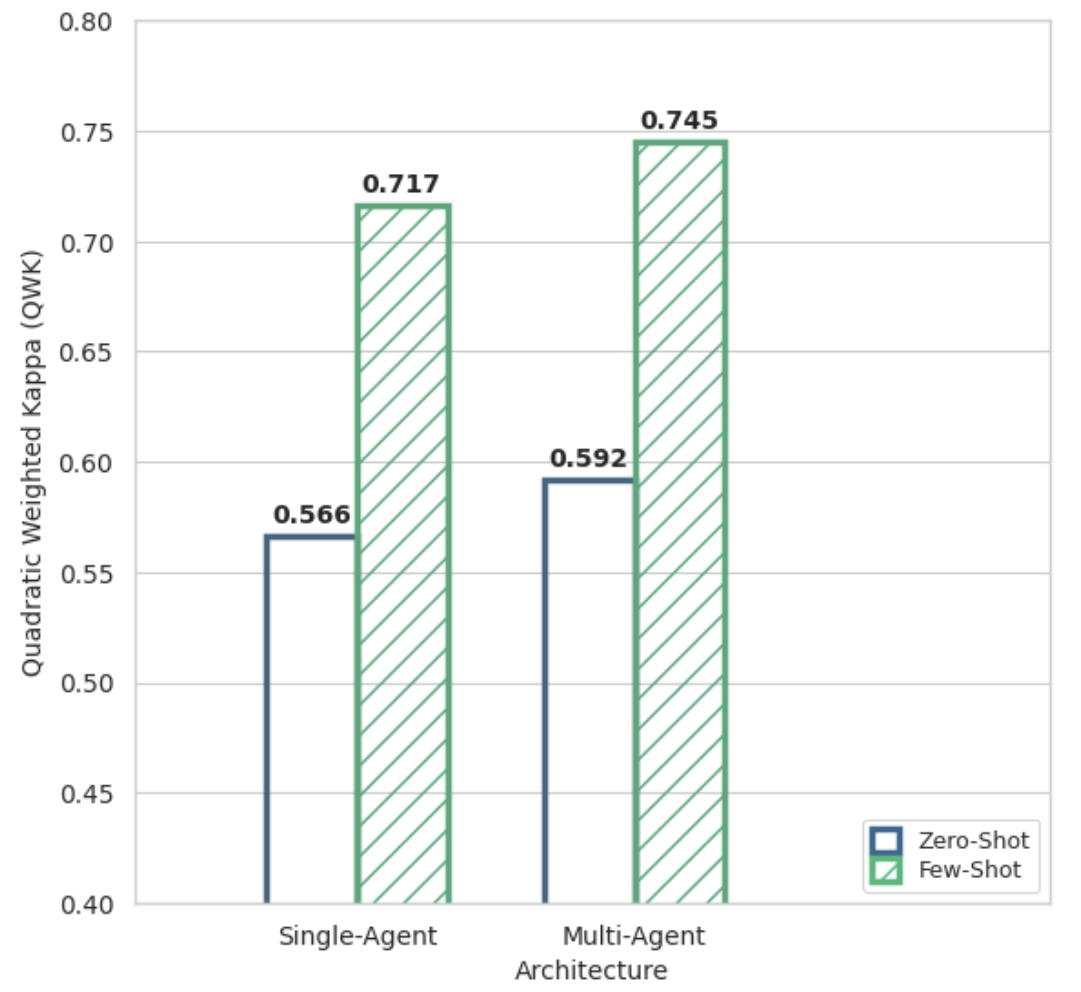}
    \caption{Performance comparison: QWK Score}
    \label{fig:3}
\end{figure}

As shown in Figure 3 and Table 1, \textbf{few-shot learning has the greatest impact on improving the performance of both architectures.} For the single-agent system, few-shot prompting improved QWK by 26.5\% (from 0.5664 to 0.7165). And for the multi-agent system, it improved QWK by 25.96\% (from 0.5917 to 0.7453). This suggests that the provision of concrete examples (2 per score level, n=12) help both architectures better calibrate their assessments.

\subsection{Performance by Score Level}
Table 2 and Figure 4 present a granular analysis of exact match accuracy across the six rubric score levels. This breakdown reveals critical insights into how different architectures handle essays at various quality levels, particularly at the distribution extremes.

\begin{table}[H]
    \centering
    \caption{Exact Match Accuracy by Rubric Score Level}
    \begin{tabular}{cccccc}
    \toprule
    Score & Total (n) & Single (Zero) & Single (Few) & Multi (Zero) & Multi (Few) \\
    \midrule
    1 & 30 & 11 (36.7\%) & 14 (46.7\%) & 17 (56.7\%) & 22 (73.3\%) \\
    2 & 120 & 67 (55.8\%) & 66 (55.0\%) & 72 (60.0\%) & 79 (65.8\%) \\
    3 & 168 & 97 (57.7\%) & 91 (54.2\%) & 87 (51.8\%) & 88 (52.4\%) \\
    4 & 103 & 44 (42.7\%) & 63 (61.2\%) & 35 (34.0\%) & 59 (57.3\%) \\
    5 & 26 & 2 (7.7\%) & 8 (30.8\%) & 3 (11.5\%) & 8 (30.8\%) \\
    6 & 3 & 0 (0.0\%) & 1 (33.3\%) & 0 (0.0\%) & 1 (33.3\%) \\
    \bottomrule
    \end{tabular}
\end{table}

\begin{figure}[H]
    \centering
    \includegraphics[width=0.9\linewidth]{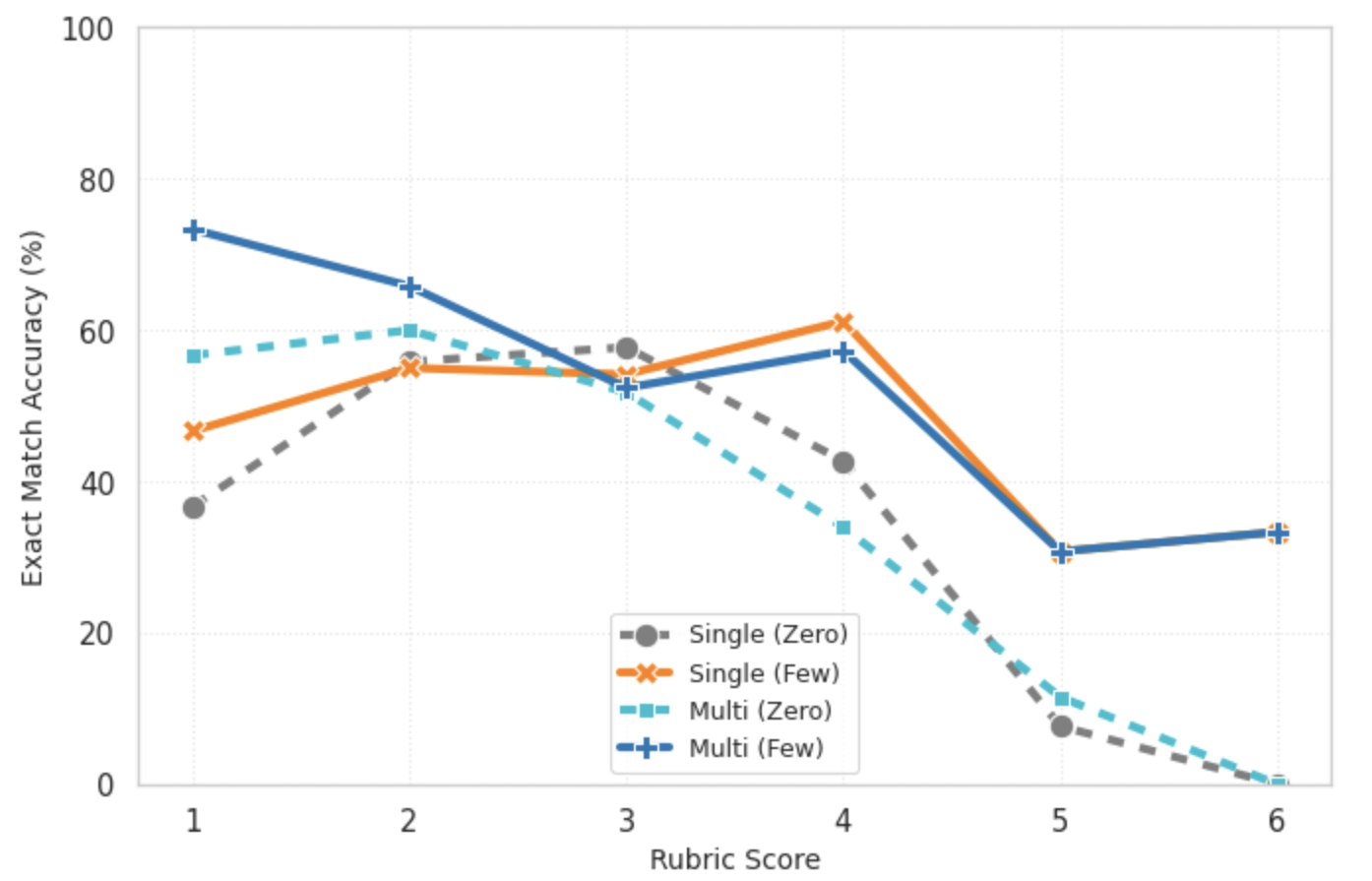}
    \caption{Accuracy across score levels}
    \label{fig:2}
\end{figure}

\subsubsection{Low-Quality Essays (Scores 1-2)}
The most striking finding from the score-level analysis is the multi-agent architecture's dramatically superior performance on essays rated 1 or 2. For score 1 essays, the multi-agent few-shot configuration achieved 73.3\% accuracy, compared to 46.7\% for the single-agent few-shot baseline, a remarkable improvement of 26.6 percentage points. Similarly, for score 2 essays, the multi-agent few-shot system achieved 65.8\% accuracy versus 55.0\% for the single-agent few-shot approach, representing a 10.8 percentage point gain.

This pattern is consistent across prompting conditions. Even in the zero-shot setting, the multi-agent architecture showed substantially better performance on low-scoring essays: 56.7\% versus 36.7\% for score 1 (a 20.0 percentage point improvement) and 60.0\% versus 55.8\% for score 2 (a 4.2 percentage point improvement).

We attribute this improved performance on low scores to the veto rule implemented in the Chairman Agent. Specifically, if any specialist agent assigns a score of 1, the final score must be 1; similarly, if any specialist assigns a 2, the final score is capped at 2 or 3. This mechanism ensures that critical deficiencies in any single dimension (content, structure, or language) appropriately constrain the overall assessment, preventing the system from over-estimating essay quality when fundamental weaknesses are present.

This finding has significant practical implications for automated essay scoring systems deployed in high-stakes educational contexts. The multi-agent architecture's ability to detect and appropriately penalize low-quality essays represents a meaningful advancement in the reliability of AES systems, particularly for identifying students who require immediate intervention.

\subsubsection{Mid-Range Essays (Scores 3-4)}
For mid-range essays (scores 3 and 4), the performance differences between architectures were less pronounced, though interesting patterns emerged. All four configurations performed similarly for score 3, ranging from 51.8\% to 57.7\% accuracy. For score 4 essays, few-shot prompting proved essential for both architectures. The single-agent few-shot configuration achieved 61.2\% accuracy compared to 42.7\% in zero-shot, while the multi-agent few-shot achieved 57.3\% versus 34.0\% in zero-shot. As with score 3, the single agent also performed better than the multi-agent for score 4.

Interestingly, the single-agent few-shot system actually achieved the highest accuracy for both scores, marginally exceeding the multi-agent few-shot system.

This pattern may reflect the inherent ambiguity in distinguishing "adequate mastery" (Score 4) from "reasonably consistent mastery" (Score 5) when using a holistic rubric. For essays in the competent-but-not-exceptional range, the characteristics that differentiate adjacent scores are subtle and often involve trade-offs across dimensions. A well-organized essay with weak vocabulary might merit a 4, while a less coherent essay with sophisticated language might also warrant a 4. The single agent's holistic view may paradoxically provide an advantage in such cases by allowing implicit compensatory scoring, balancing strengths and weaknesses in ways that specialist agents, focused on their narrow domains, cannot.

\subsubsection{High-Quality Essays (Scores 5-6)}
Both architectures struggled significantly with high-quality essays. For score 5 essays, the best-performing configurations (both few-shot variants) achieved only 30.8\% accuracy, while zero-shot configurations performed substantially worse at 7.7\% and 11.5\%. For score 6 essays, which represented only 3 essays in the test set (0.7\%), both few-shot configurations correctly identified only 1 essay (33.3\%), while zero-shot configurations failed entirely.

It is important to note that the results for score 6 are statistically insufficient to draw robust conclusions due to the extreme sparsity of the data. However, the broader trend across score 5 suggests a consistent under-prediction bias at the upper end of the distribution. Both systems frequently assigned scores of 3 or 4 to essays that human raters had marked as 5 or 6. This "regression to the mean" is likely compounded in the multi-agent architecture by its capping logic; while this mechanism effectively catches weak essays, it introduces a conservative bias where a specialist agent’s concern regarding a single dimension (e.g., a mechanical flaw) can artificially constrain the holistic score

\subsection{The Impact of Few-Shot Prompting}
Few-shot prompting emerged as a critical factor in system performance across both architectures. The provision of 12 calibration examples (2 per score level) consistently improved performance.

For the single-agent architecture, few-shot prompting produced a 26.5\% improvement in QWK and a 4.89 percentage point improvement in exact match accuracy. The benefits were most pronounced for mid-to-high range essays, with score 4 improving from 42.7\% to 61.2\% accuracy. This suggests that calibration examples help the single-agent model better understand the specific characteristics that distinguish adjacent score levels, particularly in the middle-to-upper range where differences are more subtle.

For the multi-agent architecture, few-shot prompting yielded a 25.96\% improvement in QWK and a 9.55 percentage point improvement in exact match accuracy. Notably, the improvements were most dramatic for low-quality essays, with score 1 accuracy improving from 56.7\% to 73.3\% and score 2 from 60.0\% to 65.8\%. This pattern suggests that calibration examples help specialist agents better identify critical deficiencies in their respective domains, which the Chairman Agent's veto and capping rules then translate into appropriately low overall scores. This high sensitivity of both architectures to the presence of calibration examples is corroborated by \hyperlink{Errica2025}{Errica et al. (2025)}, who emphasize that LLM consistency is often heavily dependent on prompt engineering and example selection.

\subsection{Multi-Agent vs. Single-Agent: A Comparative Analysis}
The comparison between single-agent and multi-agent architectures reveals that neither approach universally dominates; rather, each exhibits distinct strengths and weaknesses that make them suited to different assessment priorities and contexts. While the multi-agent architecture achieved marginally higher overall performance (QWK of 0.7453 vs. 0.7165 in few-shot conditions), this 2.88 percentage point improvement represents only a 4\% relative gain, a difference that may not justify the 4x computational cost increase in all deployment scenarios. This is similar to the conclusion by \hyperlink{WangDing2025}{Wang et al. (2025)}, who found that multi-agent workflows did not inherently improve classification accuracy over single-agent models in collaborative problem-solving tasks, and \hyperlink{Li2025}{Li et al. (2025)}, who demonstrated that single-agent few-shot strategies can outperform multi-agent systems in student reflection assessment. More importantly, granular analysis reveals complementary performance patterns: the multi-agent system excels at identifying critically weak essays (scores 1-2), while the single-agent system demonstrates superior or competitive performance on mid-range essays (scores 3-4). This suggests that architectural choice should be driven by specific use case requirements rather than by overall metrics alone.

The multi-agent architecture's primary strength lies in its systematic detection of fundamental deficiencies through specialist agents and aggregation rules. The veto and capping logic ensure that critical failures in any dimension (content, structure, or language) appropriately constrain the overall score, resulting in 73.3\% accuracy on score 1 essays and 65.8\% on score 2 essays, representing 26.6 and 10.8 percentage point improvements over the single-agent baseline, respectively. This architectural decomposition also offers potential interpretability advantages, as each specialist agent (content, structure, language) provides dimension-specific scores and rationales that could support targeted or personalized feedback (\hyperlink{Mansour2024}{Mansour et al., 2024}; \hyperlink{Idowu2024a}{Idowu et al., 2024}) and transparent decision-making (\hyperlink{Xie2024}{Xie et al., 2024}). However, these benefits come with significant trade-offs: the multi-agent system requires four LLM calls per essay (increasing API costs and latency by 4x) and exhibits conservative bias that constrains performance on high quality essays.

Conversely, the single-agent architecture offers computational efficiency, simpler implementation, and holistic assessment approach. However, it has strong blind spots in identifying students who require immediate intervention which is one of the most critical needs for authorities. The absence of dimension-specific assessments also limits interpretability and diagnostic value, though this may be less critical in contexts where formative feedback is not the primary goal. Table 3 summarizes these architectural trade-offs across key evaluation dimensions.

\begin{table}[H]
    \centering
    \caption{Comparative Analysis of Single-Agent vs. Multi-Agent Architectures}
    \begin{tabular}{>{\raggedright\arraybackslash}p{3.5cm} >{\raggedright\arraybackslash}p{3cm} >{\raggedright\arraybackslash}p{3.5cm} >{\raggedright\arraybackslash}p{4cm}}
    \toprule
    \textbf{Dimension} & \textbf{Single-Agent} & \textbf{Multi-Agent} & \textbf{Implication} \\
    \midrule
    Overall Performance (QWK) & 0.7165 & 0.7453 & Marginal difference; multi-agent slightly better \\
    \arrayrulecolor{lightgray}\midrule\arrayrulecolor{black}
    Low-Quality Essays (Scores 1-2) & Weaker (36.7-55.0\%) & Stronger (65.8-73.3\%) & Multi-agent better at identifying critical deficiencies \\
    \arrayrulecolor{lightgray}\midrule\arrayrulecolor{black}
    Mid-Range Essays (Scores 3-4) & Stronger (54.2-61.2\%) & Weaker/Comparable (52.4-57.3\%) & Single-agent better handles nuanced quality trade-offs \\
    \arrayrulecolor{lightgray}\midrule\arrayrulecolor{black}
    High-Quality Essays (Scores 5-6) & Weak (7.7-33.3\%) & Weak (11.5-33.3\%) & Both struggle; no clear advantage \\
    \arrayrulecolor{lightgray}\midrule\arrayrulecolor{black}
    Computational Cost & 1x (single API call) & 4x (four API calls) & Single-agent significantly more efficient \\
    \arrayrulecolor{lightgray}\midrule\arrayrulecolor{black}
    Calibration/Few-shot Dependency & High - 26.5\% improvement & High 25.96\% improvement & Both significantly depend on calibration examples \\
    \arrayrulecolor{lightgray}\midrule\arrayrulecolor{black}
    Zero-Shot Performance & QWK 0.5664 & QWK 0.5917 & Minimal difference without calibration \\
    \arrayrulecolor{lightgray}\midrule\arrayrulecolor{black}
    Interpretability & Limited (holistic only) & Enhanced (dimension-specific) & Multi-agent offers diagnostic potential \\
    \arrayrulecolor{lightgray}\midrule\arrayrulecolor{black}
    Decision Transparency & Black-box & Rule-based aggregation & Multi-agent more auditable \\
    \arrayrulecolor{lightgray}\midrule\arrayrulecolor{black}
    Optimal Use Case & High-volume, cost-sensitive & Early intervention screening & Different architectures suit different needs \\
    \bottomrule
    \end{tabular}
\end{table}

\section{Conclusion}

\subsection{Key Contributions}
This study makes three primary contributions to the field of automated essay scoring and multi-agent LLM systems:
\begin{itemize}
    \item We provide an empirical comparison of single-agent versus multi-agent architectures that represent a different research question than prior work. We demonstrate that architectural choice fundamentally shapes performance patterns in ways that aggregate metrics obscure. While prior research prioritized maximizing overall correlation with human scores, our granular analysis reveals differential effects across the score levels.
    \item We introduce a novel multi-agent design that combines domain decomposition (Content, Structure, Language specialists) with logic-based aggregation rules (veto and capping mechanisms). Unlike approaches relying on simple voting or averaging, our Chairman Agent implements rubric-aligned decision logic, ensuring that critical deficiencies in any single dimension appropriately constrain the overall score.
    \item We provide granular performance analysis that reveals systematic strengths and weaknesses of each architecture across the quality spectrum, accompanied by practical guidance for deployment contexts. Our findings that few-shot calibration is essential for both architectures, that multi-agent systems incur 4x computational costs for marginal overall gains, and that both architectures struggle with exceptional essays (scores 5-6) offer actionable insights for practitioners. By documenting not just what works but where and why different approaches excel or fail, this study provides an empirical foundation for evidence-based architectural selection in educational technology.
\end{itemize}

\subsection{Implications for Educational Practice}
The results of this study carry several implications for the practical deployment of LLM-based essay grading systems in educational contexts. First, the superior performance of the multi-agent architecture on low-quality essays suggests that such systems may be particularly valuable as screening tools for identifying at-risk students who require immediate intervention. Second, the critical importance of few-shot calibration examples highlights a practical implementation consideration: organizations deploying such systems must invest in curating high-quality, representative examples across the score range. The substantial performance gains from just 2 examples per score level suggest that this investment, while modest, is essential for system effectiveness. Third, the interpretability benefits of the multi-agent architecture, through separate content, structure, and language assessments, could improve the formative value of automated feedback. Rather than receiving only a holistic score, students could receive dimension-specific guidance on where to focus their revision efforts, potentially supporting learning and improvement more effectively than single-score systems.

\subsection{Limitations and Future Research Directions}
Several limitations of this study warrant acknowledgment and suggest directions for future research. First, the study was conducted on a single dataset (ASAP 2.0) using essays written to a limited set of prompts by U.S. secondary students. Generalizability to other writing contexts, genres, or student populations remains an open question. Future research should evaluate these architectural approaches across diverse datasets. Also, while we tested zero-shot and few-shot conditions, we did not systematically vary the number of calibration examples or explore other prompting strategies that might enhance performance. Research examining the optimal number and selection criteria for calibration examples could provide practical guidance for system deployment. Similarly, investigation of advanced prompting techniques such as chain-of-thought reasoning or self-consistency methods might further improve performance.

In addition, while we proposed that the multi-agent architecture offers interpretability advantages through dimension-specific scores and rationales, we did not empirically evaluate the quality or usefulness of these explanations. Future research could include human evaluation studies examining whether the multi-agent system's feedback is more actionable, accurate, or helpful for students and educators compared to single-agent explanations.

Meanwhile, the conservative bias observed in high-score predictions raises questions about fairness and equity. Future research should examine whether certain student populations or writing styles are systematically disadvantaged by either architectural approach, and whether bias mitigation techniques in LLMs could enhance fairness without sacrificing overall performance, as reported by \hyperlink{Idowu2024b}{Idowu (2024)} regarding traditional ML.

Finally, our aggregation approach used deterministic rules (veto and capping logic) based on domain knowledge of the rubric structure. Alternative aggregation methods, such as weighted averaging, attention-based fusion, or learned aggregation functions, might yield different performance profiles and warrant investigation.

\section*{References}
\begin{hangparas}{.5cm}{1}

\hypertarget{Chu2024}{}Chu, S., Kim, J., Wong, B., \& Yi, M. (2024). Rationale behind essay scores: Enhancing S-LLM’s multi-trait essay scoring with rationale generated by LLMs. In \emph{arXiv [cs.CL]}. \url{http://arxiv.org/abs/2410.14202}

\hypertarget{Crossley2025}{}Crossley, S. A., Baffour, P., Burleigh, L., \& King, J. (2025). A large-scale corpus for assessing source-based writing quality: ASAP 2.0. \emph{Assessing Writing}, 65(100954), 100954. \url{https://doi.org/10.1016/j.asw.2025.100954}

\hypertarget{Errica2025}{}Errica, F., Siracusano, G., Sanvito, D., \& Bifulco, R. (2025). What did I do wrong? Quantifying LLMs’ sensitivity and consistency to prompt engineering. In \emph{arXiv [cs.LG]}. \url{http://arxiv.org/abs/2406.12334}

\hypertarget{Figueras2025}{}Figueras, C., Farazouli, A., Cerratto Pargman, T., McGrath, C., \& Rossitto, C. (2025). Promises and breakages of automated grading systems: a qualitative study in computer science education. \emph{Education Inquiry}, 1-22.

\hypertarget{Guo2025}{}Guo, S., Wang, Y., Yu, J., Wu, X., Ayik, B., Watts, F. M., Latif, E., Liu, N., Liu, L., \& Zhai, X. (2025). Artificial intelligence bias on English language learners in automatic scoring. In \emph{Lecture Notes in Computer Science} (pp. 268–275). Springer Nature Switzerland.

\hypertarget{Hong2025}{}Hong, S., Cai, C., Du, S., Feng, H., Liu, S., \& Fan, X. (2025, July). “My Grade is Wrong!”: A Contestable AI Framework for Interactive Feedback in Evaluating Student Essays. In \emph{International Conference on Artificial Intelligence in Education} (pp. 27-35). Cham: Springer Nature Switzerland.

\hypertarget{Idowu2024b}{}Idowu, J.A. (2024). Debiasing Education Algorithms. \emph{Int J Artif Intell Educ}, 34, 1510–1540. \url{https://doi.org/10.1007/s40593-023-00389-4}

\hypertarget{Idowu2024a}{}Idowu, J. A., Koshiyama, A. S., \& Treleaven, P. (2024). Investigating algorithmic bias in student progress monitoring. \emph{Computers and Education: Artificial Intelligence}, 7(100267), 100267. \url{https://doi.org/10.1016/j.caeai.2024.100267}

\hypertarget{Kang2026}{}Kang, X., \& Kong, F. (2026). Decomposing topic relevance: A multi-agent LLM approach for automated essay scoring and feedback. In \emph{Lecture Notes in Computer Science} (pp. 325–336). Springer Nature Singapore.

\hypertarget{Lagakis2024}{}Lagakis, P., \& Demetriadis, S. (2024). EvaAI: A multi-agent framework leveraging large language models for enhanced automated grading. In \emph{Generative Intelligence and Intelligent Tutoring Systems} (pp. 378–385). Springer Nature Switzerland.

\hypertarget{Latif2024}{}Latif, E., \& Zhai, X. (2024). Fine-tuning ChatGPT for automatic scoring. \emph{Computers and Education: Artificial Intelligence}, 6(100210), 100210. \url{https://doi.org/10.1016/j.caeai.2024.100210}

\hypertarget{Li2025}{}Li, G., Chen, L., Tang, C., Švábenský, V., Deguchi, D., Yamashita, T., \& Shimada, A. (2025). Single-agent vs. Multi-agent LLM strategies for automated student reflection assessment. In \emph{arXiv [cs.LG]}. \url{http://arxiv.org/abs/2504.05716}

\hypertarget{Mansour2024}{}Mansour, W. A., Albatarni, S., Eltanbouly, S., \& Elsayed, T. (2024, May). Can large language models automatically score proficiency of written essays?. In \emph{Proceedings of the 2024 Joint International Conference on Computational Linguistics, Language Resources and Evaluation (LREC-COLING 2024)} (pp. 2777-2786).

\hypertarget{Sessler2025}{}Seßler, K., Fürstenberg, M., Bühler, B., \& Kasneci, E. (2025, March). Can AI grade your essays? A comparative analysis of large language models and teacher ratings in multidimensional essay scoring. In \emph{Proceedings of the 15th International Learning Analytics and Knowledge Conference} (pp. 462-472).

\hypertarget{Su2025}{}Su, J., Yan, Y., Gao, Z., Zhang, H., Liu, X., \& Hu, X. (2025). CAFES: A collaborative multi-agent framework for multi-granular multimodal Essay Scoring. In \emph{arXiv [cs.CL]}. \url{http://arxiv.org/abs/2505.13965}

\hypertarget{WangDing2025}{}Wang, Y., Ding, Z., Wu, X., Sun, S., Liu, N., \& Zhai, X. (2025). AutoSCORE: Enhancing automated scoring with multi-agent large language models via Structured COmponent REcognition. In \emph{arXiv [cs.CL]}. \url{http://arxiv.org/abs/2509.21910}

\hypertarget{WangGopal2025}{}Wang, Yu, Gopalakrishnan, M., \& Bergner, Y. (2025). Using generated rubrics to provide a window into item evaluation with multi-agent LLMs. In \emph{Lecture Notes in Computer Science} (pp. 203–217). Springer Nature Switzerland.

\hypertarget{Xie2024}{}Xie, W., Niu, J., Xue, C. J., \& Guan, N. (2024). Grade like a human: Rethinking automated assessment with large language models. In \emph{arXiv [cs.AI]}. \url{http://arxiv.org/abs/2405.19694}

\end{hangparas}

\end{document}